# Multi-view reconstruction of bullet time effect based on improved NSFF model

1st Linquan Yu
College of Automation
Beijing Institute of Technology
*Beijing, China*
ylq1959671756@163.com

2nd Yan Gao
College of Automation
Beijing Institute of Technology
*Beijing, China*
gaoem01@bit.edu.cn

3rd Yangtian Yan
College of Automation
Beijing Institute of Technology
*Beijing, China*
yyt3220200816@163.com

4th Wentao Zeng
College of Automation
Beijing Institute of Technology
*Beijing, China*
zengwentao613@163.com

*Abstract*—Bullet time is a type of visual effect commonly used in film, television and games that makes time seem to slow down or stop while still preserving dynamic details in the scene. It usually requires multiple sets of cameras to move slowly with the subject and is synthesized using post-production techniques, which is costly and one-time. The dynamic scene perspective reconstruction technology based on neural rendering field can be used to solve this requirement, but most of the current methods are poor in reconstruction accuracy due to the blurred input image and overfitting of dynamic and static regions. Based on the NSFF algorithm, this paper reconstructed the common time special effects scenes in movies and television from a new perspective. To improve the accuracy of the reconstructed images, fuzzy kernel was added to the network for reconstruction and analysis of the fuzzy process, and the clear perspective after analysis was input into the NSFF to improve the accuracy. By using the optical flow prediction information to suppress the dynamic network timely, the network is forced to improve the reconstruction effect of dynamic and static networks independently, and the ability to understand and reconstruct dynamic and static scenes is improved. To solve the overfitting problem of dynamic and static scenes, a new dynamic and static cross entropy loss is designed. Experimental results show that compared with original NSFF and other new perspective reconstruction algorithms of dynamic scenes, the improved NSFF-RFCT improves the reconstruction accuracy and enhances the understanding ability of dynamic and static scenes.

## I. INTRODUCTION

Bullet time effects are when an action or scene is slowed down and shot at different angles and perspectives. This usually requires setting up several high-speed cameras on the shooting site to shoot different angles and perspectives of the same scene, and analyzing and rendering each frame shot by computer technology in post-production. While it's a disruptive visual experience, the technology is expensive because of the need to set up multiple cameras. Therefore, low-cost dynamic scene reconstruction technology is of great significance for film and television production.

With the rapid development of neural rendering field technology, an important breakthrough has been made in new perspective reconstruction of scenes based on monocular camera. The neural rendering technology represented by Nerf[1] has greatly improved the effect of new perspective reconstruction of monocular static scenes. Compared with the traditional rendering technology, which requires multiple shots of different angles or the use of complex equipment such as 3D scanner for scene data acquisition, the data acquisition method of neural rendering technology is simpler. It only requires the use of ordinary cameras or video cameras to capture a single scene, and enough training data can be obtained with low collection cost. In addition, traditional geometric-based rendering techniques require multiple shots and processing of the scene, and Nerf can output rendering results from any perspective by shooting the scene at different angles and feeding the images and corresponding depth maps into the neural network for training. In contrast, it has higher data acquisition efficiency. However, this kind of model can only reconstruct the scene containing only static objects, and cannot render the scene containing dynamic objects (such as human body, vehicles, etc.) from a new perspective. With the development of neural deformation field and optical flow methods, various Nerf-based methods have emerged in the field of dynamic scene reconstruction and rendering from new perspectives.

In this study, the bullet special effects scenes in film and television works will be reconstructed from a new perspective based on neural rendering method. Currently, the reconstruction accuracy of dynamic scenes based on neural rendering field still needs to be improved for two main reasons. First, the neural rendering accuracy of dynamic scenes is affected by the shooting accuracy, and the shooting ambiguity caused by the movement of objects in the scene will lead to inaccurate reconstruction results. Second, the separation effect of dynamic and static scenes is not good. When the scene contains static background and dynamic objects, neural rendering may not be able to separate the two parts well, resulting in some inaccurate reconstruction results. In this study, the neural rendering field was improved to improve the reconstruction accuracy of bullet time effect.

## II. RELATED WORK

### A. Neural Radiance Fields

Nerf[1] is a 3D scene reconstruction method based on neural rendering fields. Nerf represents 3D scenes as neural rendering fields and uses a deep neural network to predict the colors and reflectance at any point in the scene. The input of the neural network is the 3D point coordinates and sight direction in the scene, and the output is the color and reflectivity of the point.

NeRF++[2] proposed by Zhang K et al. is an improvement on NeRF, mainly adding a variety of rendering techniques, such as multi-scale NeRF and progressive training. In addition, NeRF++ uses gradient clipping and depth normalization technology to improve model stability and normalization of rendering quality. Yu A et al. proposed a 3D rendering method based on neural networks[3], which can transform the rendering process in NeRF into a pixel-level image synthesis process. PixelNeRF can achieve faster and higher quality rendering by combining NeRF's rendering process with convolutional neural networks. MipNeRF[4] proposed by Barron J T et al can greatly reduce computing costs and speed up rendering while maintaining high quality rendering by introducing multi-resolution representation and layered rendering technology. NSVF[5] proposed by Liu L et al is a neural rendering method based on sparse voxels, which can represent 3D scenes as a sparse neural network model. NSVF allows for higher quality rendering and better scene reconstruction.

*B. Dynamic scene reconstruction*

The original neural radiation field [1] can only be used to represent static scenes and objects, and cannot reconstruct moving objects in dynamic scenes. To solve this problem, many researchers have proposed different methods to deal with dynamic changes, which allow new viewpoint synthesis of dynamic scenes, and can be widely used in film and television production and other fields.

D-NeRF[6], proposed by Pumarola A et al., is an earlier dynamic scene reconstruction algorithm based on neural rendering technology. By introducing deformable mesh, the deformable objects in the scene are modeled as continuous deformable flow. D-NeRF models and predicts the motion of objects in the input video, enabling high-quality reconstruction of dynamic scenes. Du Y et al. [7] 's NeRFlow simulates deformation with infinitesimal displacements and requires integration with Neural ODE[8] to obtain offsets. Park et al. have proposed Nerfies[9] for the reconstruction of character selfie scenes, namely, free-perspective selfies. Nerfies automatically decodes the underlying code for the morphing and appearance of each input view and learns the low-frequency components first, avoiding local minimums by over-fitting high-frequency details. HyperNeRF[10], proposed by Park K et al., is an extension of Nerfies [9], using a typical hyperspace rather than a single typical framework, which allows for handling scenarios with topological variations. NSFF[11] proposed by Mildenhall et al is a new perspective reconstruction algorithm of dynamic scene based on neural rendering technology. The algorithm uses the concept of flow field to model the motion and deformation in the scene as a continuous flow field, and then uses neural network to reconstruct the scene and render it from a new perspective.

*C. Deblur Nerf*

Traditional image deblurring algorithms are usually based on image degradation models, such as blind fuzzy algorithm, non-blind fuzzy algorithm and so on. In recent years, image deblurring algorithms based on deep learning have gradually become a research hotspot, among which common algorithms include algorithms based on convolutional neural networks, such as DeblurGAN[12] and SRN[13], and end-to-end neural network algorithms, such as DeepDeblur[14] and UDVD[15].

Deblur-NeRF[16], proposed by Jiawei Zhang et al., in 2021, is an image debluration algorithm based on neural rendering technology. This algorithm combines NeRF[1] and image deblurring technology, and reconstructs and renders fuzzy images by training neural networks, so as to achieve high quality deblurring and new perspective rendering. The main innovation of Deblur-NeRF lies in the introduction of image deblurring technology, which enables the algorithm to reconstruct and render images more truly and clearly. The algorithm uses the deep learning model to estimate the fuzzy kernel, and considers the influence of the fuzzy kernel in the reconstruction process, so as to realize the deblurring of the fuzzy image.

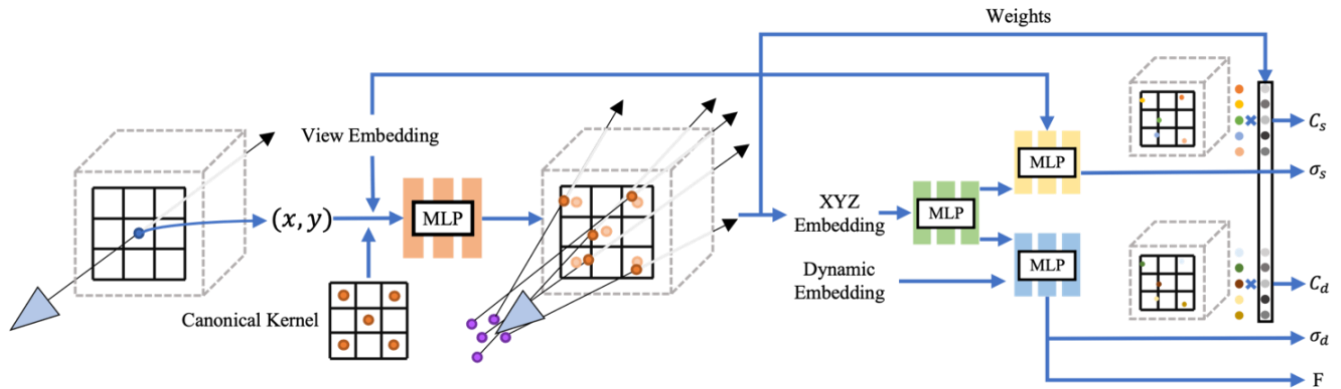

Fig. 1. Overall network structure diagram

III. IMPROVEMENT DETAILS OF NSFF

*A. Original NSFF model*

The main idea of NSFF is to try to make the model actively distinguish between dynamic scenes and static scenes by optical flow method, and adopt different modeling strategies respectively. For static scenes, the processing method of NSFF is similar to that of traditional neural rendering algorithm, that is, in the training stage, each perspective in the static scene is reconstructed, and then the reconstructed results are used to render the new perspective. Since the objects in the static scene do not deform, the same neural network can be used to reconstruct all views without considering the factor of time.

The dynamic scene processing method based on optical flow in NSFF uses a loss function of optical flow consistency, that is, the information of optical flow field is added to the training of neural renderer, and the rendering result is

compared with the real image. Specifically, it uses the following loss functions:

- Optical flow consistency loss

  This loss is based on the assumption of no distortion in time, that is, the color rendering values of the sampled points on the ray at time $i$ are equal to the color rendering values of these points at the next moment. This assumption assumes that the light flow direction is small, and the formula are as follows:

  $$\widehat{C}_{j \to i}(r_i) = \int_{t_n}^{t_f} T_j(t)\sigma_j\left(r_{i \to j}(t)\right) c_j(r_{i \to j}(t), d_i) dt$$
  $$\text{, where } r_{i \to j}(t) = r_i(t) + f_{i \to j}(r_i(t)) \quad (1)$$

  $$L_{pho} = \sum_{r_i} \sum_{j \in N(i)} \|\widehat{C}_{j \to i}(r_i) - C_i(r_i)\|_2^2 \quad (2)$$

  $r, t, \sigma, c$ represent the ray index of space, time, color, transparency, and predicted, $\widehat{C}_{j \to i}(r_i)$ said ray $j$ moment $r_i$ according to position and optical flow computation time sampling points on the pixel value. $f_{i \to j}$ represents the light flow direction from time $i$ to time $j$.

- Optical circulation loss

  This loss is considered that the optical flow quantity before and after a pixel point in two adjacent frames of images predicted by the network is a vector of equal magnitude and opposite direction. The formula is as follows:

  $$\mathcal{L}_{cyc} = \sum_{x_i} \sum_{j \in i \pm 1} w_{i \to j} \|f_{i \to j}(x_i) + f_{j \to i}(x_{i \to j})\|_1 \quad (3)$$

  Where **x** represents the spatial position vector of sampling points after position coding.

- Geometric prior loss

  The original NSFF added two prior losses, depth loss and optical flow loss, into the network, in order to make the network have better spatial perception and position prediction ability. The formula is as follows:

  $$\mathcal{L}_{geo} = \sum_{r_i} \sum_{j \in \{i \pm 1\}} \| \widehat{p}_{i \to j}(r_i) - p_{i \to j}(r_i) \|_i \quad (4)$$

  $$\mathcal{L}_z = \sum_{r_i} \|\widehat{Z}_i^*(r_i) - Z_i^*(r_i)\|_1 \quad (5)$$

  $$\mathcal{L}_{data} = \mathcal{L}_{geo} + \beta_z \mathcal{L}_z \quad (6)$$

  Where, $\mathcal{L}_{data}$ is the prior loss after combination, $\mathcal{L}_{geo}$ and $\mathcal{L}_z$ are depth loss and optical flow loss, respectively, and $\beta_z$ is the weight of optical flow fusion. $\widehat{Z}_i^*(r_i)$ is a ray, sampling depth values are weighted in the depth of the two-dimensional plane, after $Z_i^*(r_i)$ [17] the depth of the calculated value; $\widehat{p}_{i \to j}(r_i)$ is a network on the ray optical flow value of the output after projection in 2 d plane optical flow forecast, $p_{i \to j}(r_i)$ for optical flow prediction result of [18].

- Reconstruct losses

$$\widehat{C}_i(r_i) = \int_{t_n}^{t_f} T_i(t)\sigma_i(t)c_i(t) dt \quad (7)$$

$$\mathcal{L}_{cb} = \sum_{r_i} \|\widehat{C}_i(r_i) - C_i(r_i)\|_2^2 \quad (8)$$

Where $\widehat{C}_i(r_i)$ and $C_i(r_i)$ respectively is ray $r_i$ forecast of color value and real value, ultimately NSFF loss function formula is:

$$\mathcal{L}_{NSFF} = \mathcal{L}_{cb} + \mathcal{L}_{pho} + \beta_{cyc} \mathcal{L}_{cyc} + \beta_{data} \mathcal{L}_{data} \quad (9)$$

*B. Deblur-model of NSFF*

By default, the input images of NSFF are clear images from different perspectives, and the input clear images are taken as true values to optimize the network reconstructed images from this perspective, so as to improve the prediction accuracy of color and transparency of 3D sampling points. The NSFF works well when these images are well captured and calibrated, but it produces noticeable artifacts when the training image is blurred. For example, when capturing a low-light scene using a long exposure setting, the image is more sensitive to camera shake, resulting in blurred camera motion. In addition, defocusing blurring is inevitable when large apertures are used to capture scenes with large depth variations, and these blurring will significantly degrade the quality of NSFF reconstruction.

To solve this problem, the fuzzy check in Deblur-nerf was introduced into the NSFF input terminal to carry out explicit modeling of the fuzzy caused by camera movement or focal length change. The frame diagram of the entire model and the position of the fuzzy kernel were shown in Fig. 1. The core idea is to explicitly model the fuzzy process by adding additional fuzzy kernel, and seek to jointly optimize clear NSFF and fuzzy parameters so that the resolved fuzzy image matches the clear input view image. The specific working process of the model is as follows: In order to make the fuzzy kernel correctly imitate the generation process of fuzzy, we first use the deformable sparse kernel module to generate multiple optimization rays during the training process, and then extend the original single rays to these generated optimization rays as the input of the improved NSFF. The render results of these rays are then mixed with different ray results corresponding to the original rays to obtain the final blur color, which is then supervised trained with the original blur input. Since the fuzzy kernel has been able to correctly simulate the generation of fuzzy process after training, the trained NSFF at this time can directly accept clear perspective input, and can be output directly by the improved NSFF without going through the fuzzy kernel during reasoning.

Using the entire fuzzy kernel for simulation at each ray would result in a dramatic increase in computing and memory costs. Therefore, in this study, the sampling Deblur-nerf[16] strategy was used to sparse the fuzzy kernel, and the superparameter N(0<N<8) was set to represent the number of simulated rays of the fuzzy process. The fuzzy kernel is used to simultaneously predict the direction offset of each ray and the offset of the origin of the camera, respectively, to simulate the blur caused by camera movement and focal length change in the shooting process. The schematic diagram is shown in the figure. The weight of different optimized rays is predicted while the offset is predicted, which is used to guide the subsequent color mixing:

$$(\Delta o_q, \Delta q, w_q) = G_\Phi(p, q', l), q' \in \mathcal{N}'(p) \quad (10)$$

$$r_q = (o + \Delta o_q) + t d_q, q = q' + \Delta q \quad (11)$$

Where **p** is the original position, **q′** represents different optimized rays on the sparse fuzzy kernel, **l** is the current observation direction, $\mathcal{N}'$ is the sparse fuzzy kernel, $\Delta \mathbf{q}$ and $\Delta \mathbf{o_q}$ are the offset from the origin after optimization, $\mathbf{r_q}$ is the offset ray formula, and $w_\mathbf{q}$ is the fusion weight of the corresponding rays. After subsequent NSFF calculation, different ray colors will be fused according to the equation (12) as the reconstructed value of the final output of the model。

$$b_p = \sum_{q \in \mathcal{N}(p)} w_q c_q, \text{w.r.t.} \sum_{q \in \mathcal{N}(p)} w_q = 1 \quad (12)$$

Since the network architecture in this study involves the joint optimization of two modules, the output scale of fuzzy kernel is reduced by 100 times to limit the fuzzy scale of fuzzy kernel actively in the training process, and each ray is initialized near the original ray to avoid the abnormal deformation of fuzzy kernel and NSFF simultaneously. In addition, One of the optimized rays is forcibly restricted near the original ray position by adding alignment losses, as follows：

$$\mathcal{L}_{ali} = \|q_0 - p\|_2 + \lambda_o \|\Delta o_{q_0}\|_2 \quad (13)$$

$\mathbf{q}_0$ and $\Delta \mathbf{o}_{\mathbf{q}_0}$ are the origin position vectors and direction vectors of the restricted rays, respectively. Considering that the fuzzy process is not reversible, we remove the optical flow cycle loss from the final loss function.

*C. Improvements to the backbone network*

Although the original NSFF divides the scene into dynamic and static parts in the prediction process, the output colors of the dynamic and static networks are fused directly after the color prediction through optical flow information. In fact, it is still equivalent to using a network to reconstruct the scene, which is easy to lead to the overfitting of the network on the training data, so the reconstruction accuracy is low in some cases.

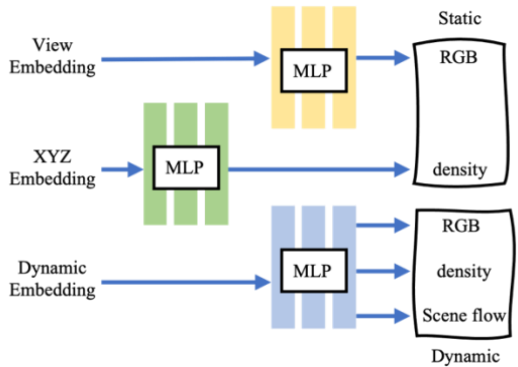

Fig. 2. Separate architecture of dynamic and static networks

This paper believes that one reason for the low accuracy of NSFF reconstruction is that there is no restriction on the prediction process of the model for dynamic and static scenes, and the goal of the network is only to reconstruct accurate fusion values rather than accurate dynamic and static outputs, so the accuracy will decline when the perspective is converted. To solve this problem, this study completely separated dynamic and static networks in NSFF, as shown in Fig. 2, and designed a training strategy for separating dynamic and static scenes. Firstly, the dynamic network is used to predict the optical flow size of three-dimensional points. The points whose value is lower than the static threshold $F_{thr}$ are regarded as static points in the scene, and the dynamic weight here is limited, as shown in the equation (14). Then the dynamic and static parts are fused to force the network to improve the prediction accuracy through the static part. In this way, the trained network has a better reconstruction ability for both static and dynamic scenes, and the result can be more accurate when the predicted color value and transparency are integrated in the final reasoning.

$$\mathcal{L}_{dy} = \frac{1}{K} \sum_{r, \|\hat{F}_r\| \leq F_{thr}} \sum_{k=1}^{K} \sigma_d^k(t) \quad (14)$$

*D. Dynamic and static cross entropy loss*

In the process of using NSFF to reconstruct dynamic scenes from a new perspective, when the test perspective differs greatly from the perspective in the training set, fog and other phenomena often occur. We believe that this phenomenon is mainly caused by the network's wrong understanding of dynamic and static scenes. For example, if there is a flag rotating 360 degrees on a flagpole, the network is likely to regard the rotating flag as a dynamic object during the reconstruction under a fixed training perspective, but the hidden flag rotating in place may be recognized by the network as a static object due to its unique motion mode. The errors that do not affect the training results of the network on the training set are actually caused by the overfitting of the network on the training set. When the original NSFF is reconstructed in a new perspective, static scenes that were previously misestimated can create fog in the reconstructed results.

This situation is avoided by adding additional prior knowledge in this study. On the same ray, firstly, the weight sum of dynamic objects within a certain range of ε is calculated by sliding the window, and then the weight sum of dynamic objects at each sampling point is calculated with the cross entropy of the corresponding static objects. The formula is as follows:

$$\mathcal{L}_{Ds} = \sum_{x_i} W_i^D \cdot \log(W_i^s) \quad (15)$$

$$W_i^D = \sum_{x_j, |j-i| \leq \varepsilon} W_j^d \quad (16)$$

Where $\mathbf{W}_D$ and $\mathbf{W}_s$ are the predicted values of dynamic weight and static weight at the sampling point on the ray, respectively. ε specifies the distance between the peak value of the dynamic part and the peak value of the static part, which is set as 12 in this paper.

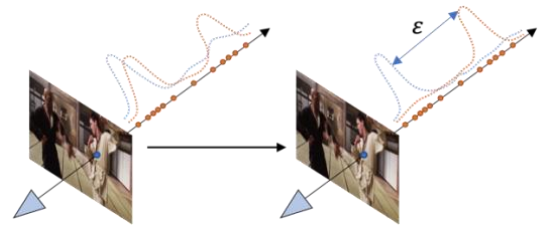

Fig. 3. The principle of dynamic and static cross entropy

The purpose of this is to encourage the network to have a certain distance between the peaks of the predicted dynamic weights and the static weights on each ray, which not only reduces the ambiguity of the network to the dynamic and static probabilities at a certain point in the scene, but also forces a certain distance between the local peaks of the dynamic objects in the space and the local peaks of the static objects, as shown in Fig. 3. It avoids the intermingling of dynamic and static regions caused by rotation, improves the coherence of dynamic and static regions, and is more in line with the real situation.

To sum up, the final loss function during training of the improved NSFF-RFCT model in this study is:

$$\mathcal{L} = \mathcal{L}_{cb} + \mathcal{L}_{pho} + \beta_1 \mathcal{L}_{data} + \beta_2 \mathcal{L}_{ali} + \beta_3 \mathcal{L}_{dy} + \beta_4 \mathcal{L}_{Ds} \quad (17)$$

## IV. EXPERIMENTS AND RESULTS

### A. Dataset and training

The experimental results of NSFF were obtained from Nvidia Dynamic Scenes Dataset[19], but the scene images in this dataset came from multiple different cameras. The Angle changes of the two adjacent frames in time were not continuous, and the shock changes of the image Angle under the time series often occurred in the training set. This is inconsistent with the continuity of perspective in the common bullet-time effects scene. In order to verify the reconstruction effect of bullet time effect achieved by the results of this study, we collected bullet time effect clips commonly seen in different movies on the Internet. People or other objects in these segments move in different degrees, and the speed of moving objects in different segments is also different, and there is only one corresponding perspective image for the scene at a certain moment in the segment. Part of the data set images are shown in Fig. 4. We divided the training set and the verification set by 9:1 according to the pictures of different scenes.

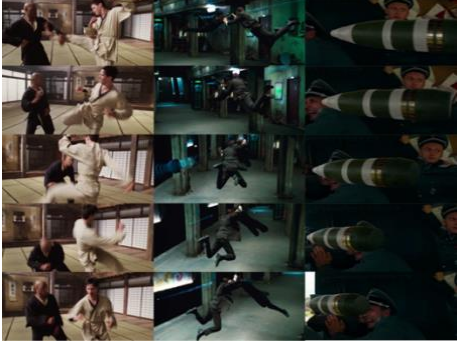

Fig. 4. Example of experimental data set

We used COLMAP for sparse reconstruction of sampled film and TV clips to obtain the estimated values of camera internal and external parameters of different frames. In order to improve the accuracy of camera pose reconstruction, we generated masks of moving objects based on category segmentation model [20], and shielded dynamic regions in the process of COLMAP feature extraction and feature point matching. Finally, we use the monocular depth prediction model [17] and RAFT[18] respectively to calculate the depth supervision map of each frame and the two-dimensional optical flow between adjacent frames of the same camera for the training of geometric prior losses in the network.

We implemented the insertion of deformable sparse kernel on NSFF based on Pytorch, and set the number of sparse positions N to 5 in Deblur-nerf[16]. No matter in the training or testing process, 128 sampling points were sampled on each ray, including 64 coarse sampling points and 64 fine sampling points respectively, and the spatial position was transformed by position coding. Before inputting the model, the scene coordinates were converted into the NDC standardized equipment coordinate space. We used two NVIDIA V100 GPUs to co-optimize the sparse core with the improved NSFF, with the ray batch set to 1024. We used the Adam optimizer with default parameters to optimize the network, and the learning rate started at 5e−4 and gradually decreased to 8e−5. It takes about 4 days for a single scene model to converge. In the test, it takes about 5s to render a single frame of 512×288.

### B. Evaluation Index

We tested three different standard error indicators of different methods in the whole scene to compare the rendering quality of different methods, namely, peak signal-to-noise ratio (PSNR), structural similarity index measure (SSIM[20]) and perceived similarity index (LPIPS[21]).

Given a $m \times n$ size of the original image and the reconstructed image, the PSNR(DB) formula of both is as follows:

$$MSE = \frac{1}{mn} \sum_{i=0}^{m-1} \sum_{j=0}^{n-1} [I(i,j) - K(i,j)]^2 \quad (18)$$

$$PSNR = 10 \cdot log_{10} \left( \frac{MAX_I^2}{MSE} \right) \quad (19)$$

Where $MAX_I^2$ is the maximum pixel value possible for the picture. For the three-channel color graph, we calculated the MSE of the three channels of RGB respectively, and then divided by 3 as the result of the equation (19).

The formula of SSIM is as follows, where $\mu_x$, $\sigma_x^2$ are the mean and variance of $x$, $\mu_y$, $\sigma_y^2$ are the mean and variance of $y$, $\sigma_{xy}$ are the covariance of $x$ and $y$, and $c_1$ and $c_2$ are two constants

$$SSIM(x,y) = \frac{(2\mu_x\mu_y + c_1)(2\sigma_{xy} + c_2)}{(\mu_x^2 + \mu_y^2 + c_1)(\sigma_x^2 + \sigma_y^2 + c_2)} \quad (20)$$

LPIPS respectively fed two images into the neural network $F$(VGG, Alexnet, Squeezenet) for feature extraction, and normalized the output of each layer after activation. Then, L2 distance was calculated after weighted by layer w, and finally the average distance was obtained。

### C. Analysis of experiment results

In order to verify the effect of our model and compare it with other models, we trained D-Nerf, Nerfies, NSFF, and our NSFF-REFT to reconstruct different scenarios in the data set under the same experimental conditions described above, and tested them using the same test set. During the training, we uniformly converted all the input pictures to $512 \times 288$ resolution. The results are shown in Table I. It can be seen that our model is more competitive than other models in the new perspective reconstruction task of bullet time effects.

TABLE I. QUANTITATIVE EVALUATION OF NOVEL VIEW SYNTHESIS

| Network model | PSNR ↑ | SSIM ↑ | LPIPS ↓ |
|---|---|---|---|
| D-Nerf | 22.13 | 0.891 | 0.062 |
| Nerfies | 28.04 | 0.909 | 0.054 |
| NSFF | 28.17 | 0.927 | 0.044 |
| NSFF-REFT | **28.34** | **0.931** | **0.042** |

For bullet time special effect scenes, our model has a better effect than the existing dynamic scene reconstruction model from new perspective. The PSNR index reaches 28.34, indicating that our model has a better reconstruction effect in details. In addition, our model achieves better effects on both SSIM and LPIPS, indicating that the reconstruction results of our model have better effects on structure and texture perception. Both Nerfies based on deformation field and NSFF based on optical flow are lower than our NSFF-REFT in various indexes.

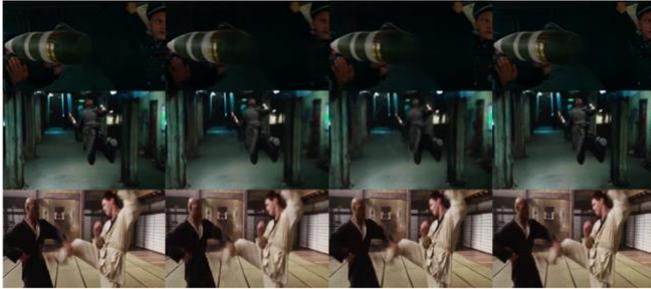

(a) ground truth  (b) Nerfies  (c) NSFF  (d) Ours

Fig. 5.  Comparison of reconstruction effects of different models

Fig. 5 shows some reconstruction results of our model and the other models mentioned above under different scenarios. It can be seen from the results that the reconstruction accuracy of our improved model is better than other models. In addition, the reconstruction result of static scene as shown in Fig. 5 is obtained after the color value and transparency are weighted only by the weight of the static area of the network. It can be seen that our modified model is better than NSFF in distinguishing between dynamic scene and static scene, reducing the degree of overfitting of the network。

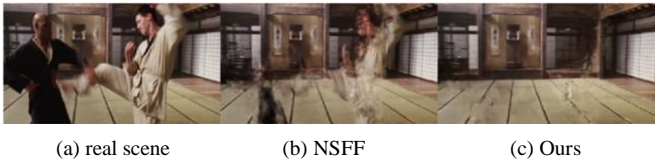

(a) real scene  (b) NSFF  (c) Ours

Fig. 6.  Dynamic scene removal effect comparison

## V. CONCLUSION

In order to solve the problem of reconstruction from new perspective of current dynamic scenes, we mainly made the following three improvements: adding DSK to the training process to explicitly model the fuzzy process; inputting the clear reconstruction sampling point information input from DSK into the rendering field to solve the reconstruction fuzzy problem caused by the common input fuzzy in dynamic scenes; In the process of color prediction, the weight of dynamic region is limited by the size of output optical flow, forcing the network to reconstruct static objects through static network, avoiding overfitting problem. A new dynamic and static cross entropy loss function is designed to make the weight peaks of the dynamic and static regions have a certain distance in the direction of rays, so as to avoid the fog problem caused by dynamic and static mixing in perspective transformation. The test results show that our model improves the accuracy of new perspective reconstruction in bullet time special effect scenes.